# HNet: Graphical Hypergeometric Networks

*Erdogan Taskesen*

**ABSTRACT**

**Motivation:** Real-world data often contain measurements with both continuous and discrete values. Despite the availability of many libraries, data sets with mixed data types require intensive pre-processing steps, and it remains a challenge to describe the relationships between variables. The data understanding phase is an important step in the data-mining process, however, without making any assumptions on the data, the search space is super-exponential in the number of variables.

**Methods:** We propose graphical hypergeometric networks (HNet), a method to test associations across variables for significance using statistical inference. The aim is to determine a network using only the significant associations in order to shed light on the complex relationships across variables. HNet processes raw unstructured data sets and outputs a network that consists of (partially) directed or undirected edges between the nodes (i.e., variables). To evaluate the accuracy of HNet, we used well known data sets and in addition generated data sets with known ground truth. The performance of HNet is compared to Bayesian structure learning.

**Results:** We demonstrate that HNet showed high accuracy and performance in the detection of node links. In the case of the Alarm data set we can demonstrate on average an MCC score of 0.33 $\pm$ 0.0002 ($P<1\times10^{-6}$), whereas Bayesian structure learning resulted in an average MCC score of 0.52 $\pm$ 0.006 ($P<1\times10^{-11}$), and randomly assigning edges resulted in a MCC score of 0.004 $\pm$ 0.0003 ($P=0.49$).

**Conclusions:** HNet can process raw unstructured data sets, allows analysis of mixed data types, it easily scales up in number of variables, and allows detailed examination of the detected associations.

**Availability**: https://erdogant.github.io/hnet/

## 1 INTRODUCTION

In recent years, much work has been done in effort to progress network-learning. The importance is indicated by the many applications that have been developed to enable modelling of complex interaction systems such as social networks, collaboration networks or biological networks. By revealing the patterns using a network, we can better understand the organizational and structural functions of network systems. Roughly speaking, the field of network learning can be dissected into two groups; the generative and discriminative models. The challenges for generative models is to learn a network representation for an existing network (e.g., social network). Commonly used techniques are graph/knowledge embeddings that transform node-links in a low-dimensional vector, which can then be used in applications with supervised or unsupervised models. Some methods that capture the complex associations between node-links are Splitter[1], Deepwalk[2], node2vec[3] and LINE[4]. On the other hand, the challenge for discriminative models is to learn the network structure or its associations (node links) given the data set. In these cases, structured data sets are used as input into the model with the goal to determine the underlying network. The questions that are addressed using discriminative modelling comprise; does variable X (in)directly influence Y, or do X and Y have a common cause? Representations can be learned using Bayesian Network structure learning[5,6] which aims to determine the directed acyclic graphs (DAG) given the data. Bayesian learning has been successfully applied in many fields such as insurance[7], health[8], and biological networks[9]. However, the search space of all possible DAGs is super-exponential in the number of variables for which the typical scoring functions can result in a local suboptimum. This is especially the case for large data sets (e.g., with many node links to be determined) where an exhaustive search is intractable due to computational burden. It is known that Bayesian approach has NP-complete[11] which makes the analysis of very large data sets difficult[12]. Nevertheless, for small data sets, an exhaustive search for DAGs can be computed, whereas for medium data sets, the use of heuristics (e.g., hill-climbing[10]) in combination with Bayesian approaches can provide a good solution. In addition to Bayesian learning, there are also rule-based machine learning techniques (association rules) to discover co-occurrence relationships between variables (item sets) in the search space. The use of rule-based techniques, such as Apriori[13], Eclat[14] and FP-Growth, has been successfully applied in many fields such as marketing (e.g., product placements, promotional pricing), retail (e.g., loyalty programs, sales promotions), security (e.g., intrusion detection[15], malicious activities), and web usage mining (e.g., advertisements). A drawback is however the risk of finding many spurious associations, and the limitation of only modelling discrete values (item lists).

With HNet, we developed a discriminative model aiming to discover statistically significant associations in data sets with mixed data types, i.e., discrete and continuous variables. The edges of the network are formed by the significant associations after applying the Holm correction for multiple testing. Furthermore, HNet does not force variables into static item sets but instead variable item sets were created to allow deep examination by the interactive network. In order to test the accuracy and performance of HNet, the detection of networks with known ground truth is evaluated (i.e., Sprinkler, and Alarm[16] data set). In addition, the Titanic[17] data set is analysed to showcase the goodness of fit, the ease of use and how to deeper examine the discovered associations.

## 2 MATERIALS AND METHODS

**HNet.** To detect significant edge probabilities between pairs of vertices (node-links) given a data set, a multi-step process is developed (Figure 1, A-F). The first step is pre-processing the data set by feature typing (Figure 1A). In this step we type each feature as discrete or numeric. Features can be excluded on user defined input parameters, such as the restriction on the minimum number of samples ($y_{min}$). The typing of features is automatically determined or can be user-defined. For automatic typing, features are set to numerical if values are of the floating kind or show more than a minimum number of unique elements (e.g., if the number of unique elements $\geq$20% of the total non-missing). Features are set to discrete if values are boolean, integer or string. The second step (Figure 1B) is encoding the discrete values into a one-hot dense matrix (dummy coding). The one-hot dense matrix is subsequently used to create combinatory features using *k* combinations over *n* features (without replacement, Figure 1C). The default for *k* is set to 1, where the input matrix with discrete matrix ($X_d$) is equal to the combinatory matrix ($X_c$). When k>1, *n* boolean features are combined by multiplication of *k* unique combinations (eq.1). Each new combinatoric feature ($X_c$) is then added to the dense matrix. To avoid high computational costs, mutual exclusive features are excluded, and features are excluded for which $X_c$ contains less then $y_{min}$ positive samples (the default is set to $y_{min}$=10, eq.2).

$$X_c = \prod_{k=1..n}^{\binom{n}{k}} (X_{c1}, X_c, .., X_N)$$

(1)

$$\sum (X_c) \geq y_{min}$$

(2)

The final dense matrix ($X_c$) is then used to assess significance with the discrete feature ($X_d$) (Figure 1D). Significance is tested using the hypergeometric distribution, where we test for over-representation of successes in feature $X_d$. The hypergeometric *P*-value between feature $X_d$ and feature $X_c$, is calculated as the probability of randomly drawing *x* or more successes from the population in *n* total draws with population size *N* containing *K* successes. For any $X_d$ and $X_c$, $P_{dc}(Xd, Xc)$ is computed as depicted in eq.3.

$$P_{dc}(X_d, X_c) = F(X \geq x|M,K,n) = 1 - \sum_{0}^{x-1} \frac{\binom{K}{x}\binom{N-K}{n-x}}{\binom{N}{n}}$$

(3)

To assess significance across the numeric features ($X_n$) in relation to the dense matrix ($X_c$) we utilized the Mann-Whitney U test. Each numeric vector $X_n$, is split on discrete feature $X_c$ versus $\sim X_c$, and then tested for significance. All tested edge probabilities between pairs of vertices, either discrete-discrete or discrete-numeric, are stored in an adjacency matrix ($P_{adj}$), and are corrected for multiple testing. The default Multiple Test Method (MTM) is set to Holm[20] (Figure 1E, equation 4) but can be set to various False Discovery Rate (FDR)[18] or Familywise error rate (FWER)[19] methods.

$$P^*_{adj} = MTM(P_{adj})$$

(4)

The last step in HNet (Figure 1F) is declaring significance for node-links. An edge is called significant under alpha is 0.05 by default. The edge-weight is computed as depicted in equation 5.

$$Edge\_weight = -log10(P^*_{adj})$$

(5)

The final output of HNet is an adjacency matrix containing edge weights that depicts the strength of pairs of vertices. The adjacency matrix can then be examined as a network representation.

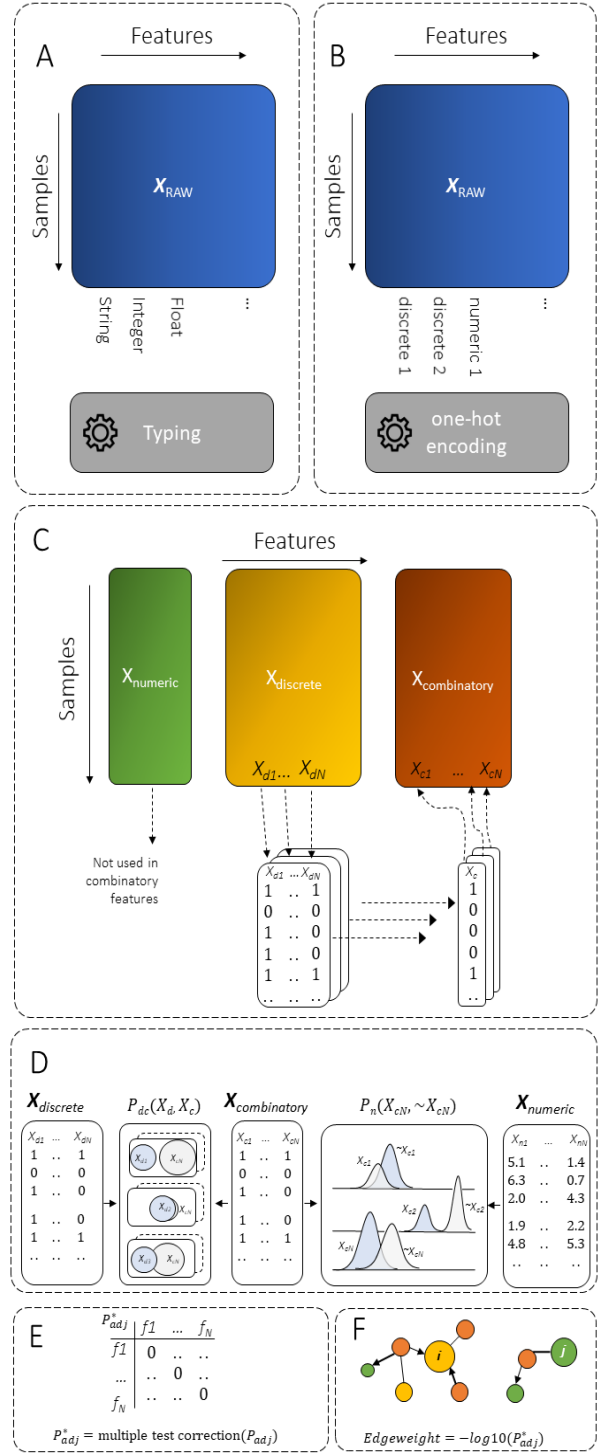

*Figure 1. Schematic overview of HNet.* To detect significant edge probabilities between pairs of vertices given a data set, a multi-step process is developed. *(A)* Each feature is set as discrete or numeric variable. *(B)* One-hot encoding or dummy coding is the transformation of discrete values into a one-hot dense matrix. *(C)* The one-hot dense matrix is used to create combinatory features using k combinations over n features without replacement. *(D)* The final dense matrix (Xc) is used to assess significance with the discrete features (Xd), and numeric features (Xn). *(E)* All tested edge probabilities between pairs of vertices, either discrete-discrete or discrete-numeric, are stored in an adjacency matrix (Padj), and are corrected for multiple testing. *(F)* The final adjacency matrix contains edge-links and edge-weights that can be represented as a network.mini/Hochberg.

**d3graph** is an interactive, stand-alone, and dynamic network graph representation to deeper examine the detected associations. Just like static networks graphs, the dynamic graph consists out of nodes and edges for which sizes and colours are adjusted accordingly. The network is created with collision and charge parameters to ensure that nodes do not overlap. Each node contains a text-label, whereas the links of associated nodes can be highlighted by double clicking on the node of interest. Furthermore, each node involves a tooltip that can easily be adapted to display the underlying data. For deeper examination of the network, edges can be gradually broken on its weight using a slider. We developed d3graph as a stand-alone python library which outputs a custom java script file based on a set of user-defined or HNet parameters. The custom java script file is built on functionalities from the d3 java script library (version 3). In its simplest form, the input for d3graph is an adjacency matrix for which the elements indicate which pairs of vertices are adjacent or non-adjacent in the graph.

**Bayesian structure learning.** With the Bayesian structure learning algorithm, we learn the optimal DAG on the subset of the data containing all discrete variables. We applied a score-based approach under the assumption that the data is complete (no missing values). The score-based model selection approach consists out of two main parts, the scoring function, and the search strategy. The scoring function maps models to a numerical score based on how well the model fits the given data, whereas the search strategy enables selection of a model with optimal score across the search space of all possible models. The Bayesian information criterion (BIC) is used as the scoring function to measure the model fit. The hill-climbing algorithm is used as search strategy. We ran Bayesian structure learning on data set containing a varying set of samples to ultimately select the best scoring model.

**Software Architecture.** The HNet library is developed in Python and consists out of various independent libraries. Documentation and examples of *HNet* are available at https://erdogant.github.io/hnet/. The pre-processing step is integrated in HNet but also independently available using the *df2onehot* library: https://github.com/erdogant/df2onehot. The output of an interactive graph is available using the library *d3graph*: https://github.com/erdogant/d3graph. Bayesian structure learning is performed using the *bnlearn* library: https://erdogant.github.io/bnlearn/.

**Data**. Directed Acyclic Graphs (DAG) of Sprinkler and Alarm[16] are used to generate a data set by means Bayesian inference and forward sampling. Generated data sets vary in sample size; N=[100,1000,5000,10000], and contain discrete features (one-hot matrix). The number of nodes for the Sprinkler model contains 4 nodes, 4 arcs, and 8 parameters. The Alarm model contains 37 nodes, 46 arcs and 509 parameters. The titanic data set contains 891 samples with 12 feature columns with mixed data types.

## 3 RESULTS

We evaluated the accuracy of HNet in the detection of edge probabilities between pairs of vertices for both directed and undirected node-links. We perform experiments on two synthetic data sets with varying number of parameters, and with known ground truth. Finally, we performed an experiment on the titanic data set which is a well-known unstructured data set and representative as an real-world application.

### 3.1 Detection of node-links using the Sprinkler data.

A natural way to study the relation between nodes in a network is to analyse the presence or absence of node-links. The sprinkler data set contains four nodes and therefore ideal to demonstrate the working of HNet in inferring a network. Links between two nodes of a network can either be undirected or directed (directed edges are indicated with arrows). Notably, a directed edge does imply directionality between the two nodes whereas undirected does not. We generated data using Bayesian forward sampling using the Conditional Probability Distributions (CPDs) for the Sprinkler system as shown in Figure 2A. Each node consists of two states, either being True or False for which we sampled with N=100, 1000, 5000, and 10000 samples. The true state is commonly used as response variable whereas the false state is seen as background. In this case the false state also describes the condition of the variable and is therefore also used as response variable. This means that in this model, the four nodes are split into eight nodes, each describing a state. The results using HNet for 1000, 5000 and 10000 samples showed consistent similar detection of significant node-links ($P_{holm} \leq 0.05$, Figure 2B and C). The inferred network contains both directed and undirected edges and does represent the initial CPD. As an example, an edge detected between Wet Grass and Sprinkler is True. When the Sprinkler is on, there is also an association between sprinkler (True) and no Rain (False), and sprinkler (True) and no Clouds (False) (Figure 2B and 2C). On the other hand, when the sprinkler is off (False), an association is seen between sprinkler (False) and Cloudy (True) and between sprinkler (False) and Rain (True). When we lower the sample size to N=100 samples, we see the absence of node-links (coloured orange in Figure 2C) compared to the data set with ≥1000 samples. To examine the minimum number of samples that shows the same node-links as seen with >1000 samples, we gradually increased the number of samples from 100 towards 1000 in steps of 10. We can demonstrate convergence of number of edges after approximately 400 samples, depending on the multiple test correction (Figure 2D). In addition, we compared the exact node-link to the network with >1000 samples. We show that detected edges converge by an increasing number of samples to a network built on a large sample size.

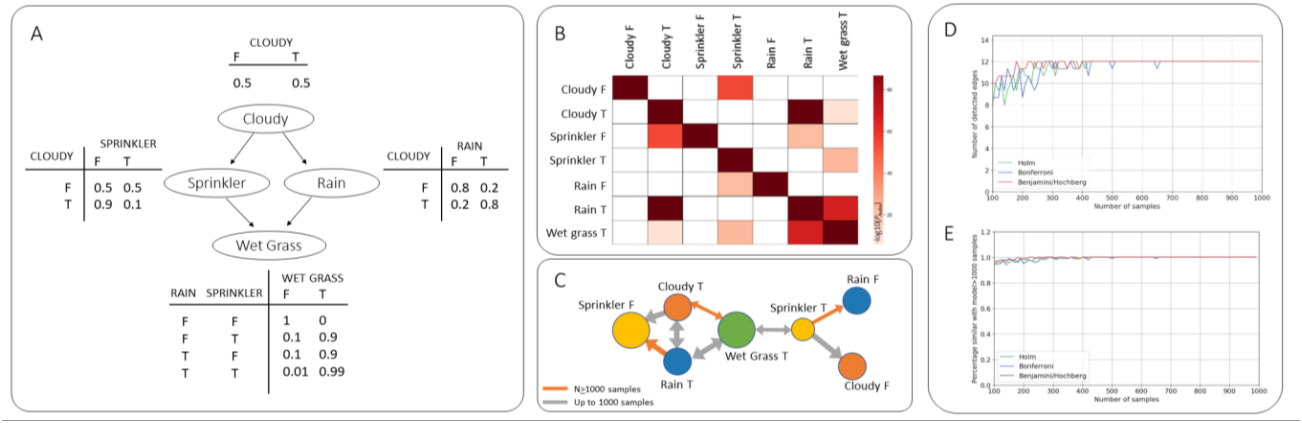

*Figure 2. Results on the Sprinkler model using HNet.* (**A**) Conditional Probability Distributions (CPD) that is used to generate data using Bayesian forward sampling. (**B**) Adjacency matrix determined by HNet with an input of N=1000 samples of the Sprinkler CPD. Elements (node-links) are coloured based on the -log10($P_{holm}$). (**C**) Graph network view of the adjacency matrix. Node size is based on the percentage of available labels. Node color is based on the unique feature names. Edge width is set by the -log10($P_{holm}$) values. The grey coloured edges are only seen up to 100 samples. For larger sample sizes ≥1000, both orange and grey edges are consistently detected. (**D**) Number of detected edges for a varying number of samples when using multiple test correction; Holm, Bonferonni or Benjamini/Hochberg. (**E**) Comparison of node-links using N=1000 samples versus a varying number of samples. The results are based on three multiple test correction methods; Holm, Bonferonni or Benjamini/Hochberg.

## 3.2 Performance of HNet compared to other methods.

To measure the performance of HNet we utilized the alarm data set, which is a medium to large network containing 37 nodes with 509 parameters. The data set is used to compare the performance of HNet with Bayesian structure learning, random results and the golden truth. To generate a data set with a ground truth, we used Bayesian forward sampling with Conditional Probability Distributions (CPDs) of the Alarm system. We sampled with N=1000, 5000 and 10.000 samples. Because some nodes consist more than two states, we only considered the true states as response variable to avoid analysing mixed background groups. Because the golden truth of node-links and edge directionality is known for the network, it can be used to examine the performance of HNet compared to Bayesian structure learning. Three experiments were set up: 1. HNet versus golden truth, 2. Bayesian structure learning versus golden truth, and 3. random adjacency matrix versus the golden truth. Each experiment is performed for the detection of directed and undirected edges (Figure 3A). The performance is measured using Matthews correlation coefficient (MCC). Note that MCC is a measure to quantify the quality of binary classifications, in this case the detected node-links and its directionality. Coefficient values range between -1 and +1 with coefficient of +1 represents a perfect prediction, 0 an average random prediction and -1 an inverse prediction. The results for N=1000, towards 10.000 samples, including edge directionality, showed an average MCC score of $0.23 \pm 0.0001$ ($P<1\times10^{-4}$) for HNet. Bayesian structure learning showed an average MCC score of $0.34 \pm 0.009$ ($P=1\times10^{-10}$), and the average MCC score when using random edges is $0.004 \pm 0.0003$ ($P=0.4$). We also analysed the specificity of the various models in the detection of undirected node-links. To make the results comparable across the various models we symmetrized the elements on the adjacency matrix with respect to the diagonal. The average MCC score for the detection of undirected edges in HNet is $0.33 \pm 0.0002$ ($P<1\times10^{-6}$), for Bayesian structure learning the average MCC score is $0.52 \pm 0.006$ ($P<1\times10^{-11}$), and the average MCC score when using random node-links is $0.004 \pm 0.0003$ ($P=0.49$). Finally, we also compared the undirected adjacency matrix of the golden truth towards the directed model which results in an average MCC score of $0.69$ ($P<1\times10^{-13}$).

## 3.3 Interpretability.

The titanic data set contains a data structure that is often seen in real use cases (i.e., the presence of discrete, boolean, and continuous variables per sample) which is therefore ideal to demonstrate the steps of HNet, and to show the interpretability. The first step is typing of the 12 input features, followed by one-hot encoding (Figure 4A). This resulted in a total of 2634 one hot encoded features for which only 18 features had the minimum required of 10 samples; Survived [1,0], Pclass [1,2,3], Sex [female, male], Sibsp [0,1,3,4], Parch [0,1,2], Cabin, Embarked [C,Q,S]. The total number of features for the model is 20, which includes the two numeric features; Fare and Age. The next step in HNet is to determine the node-links for which in total 60 unique edges across 47 nodes are detected (alpha=0.05 and multiple testing correction is Holm, Figure 2A). Note that the detected node-links can be indicative for directionality, as an example no survival (survived=0) is significantly associated with males, but not the other way around. Therefore, directionality can be seen from males to no-survival. Although the ground truth of this data set is unknown, the strongest association is in line with intuitive expectations, i.e., first class passengers are significantly associated with High Fare (fare$\geq$60.3, $P<2.87^{-79}$), whereas third class passengers are significantly associated with Low Fare (fare$\leq$8.1, $P<4.99^{-73}$). The next best association is between passengers that are female and survived ($P<4.86^{-57}$), followed by male passengers that did not survive ($P=4.79^{-57}$). The network graph is consistently expanded across the survived yes/no clusters. The male-no survival cluster is expanded with low fare, having no siblings, embarking position is S or Q whereas the females-survival cluster is expanded with high fare, having 1 sibling, and embarking position is C. Interestingly, directionality for passengers that did not survive is outwards whereas those that survived is mainly inwards. This may suggest that surviving passengers were more likely to be included in coordinated actions, whereas this was not the case for passengers that did not survive.

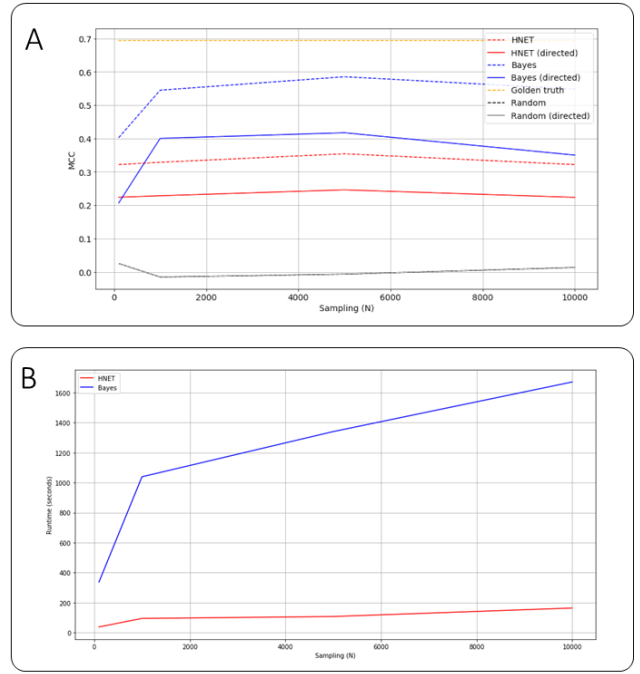

**Figure** *3. Results on Asia model using HNet. For the Asia CPD we sampled up to N=10.000 samples and compared the performance between HNet and Bayesian structure learning, random results and the golden truth. (**A**) MCC score between models for directed is shown with dashed line, and straight line for directed. Models are depicted with different colors; HNet=red, Bayesian structure learning=blue, Random=black, golden truth=yellow. (**B**) Runtime for HNet and Bayesian structure learning over increasing number of samples.*

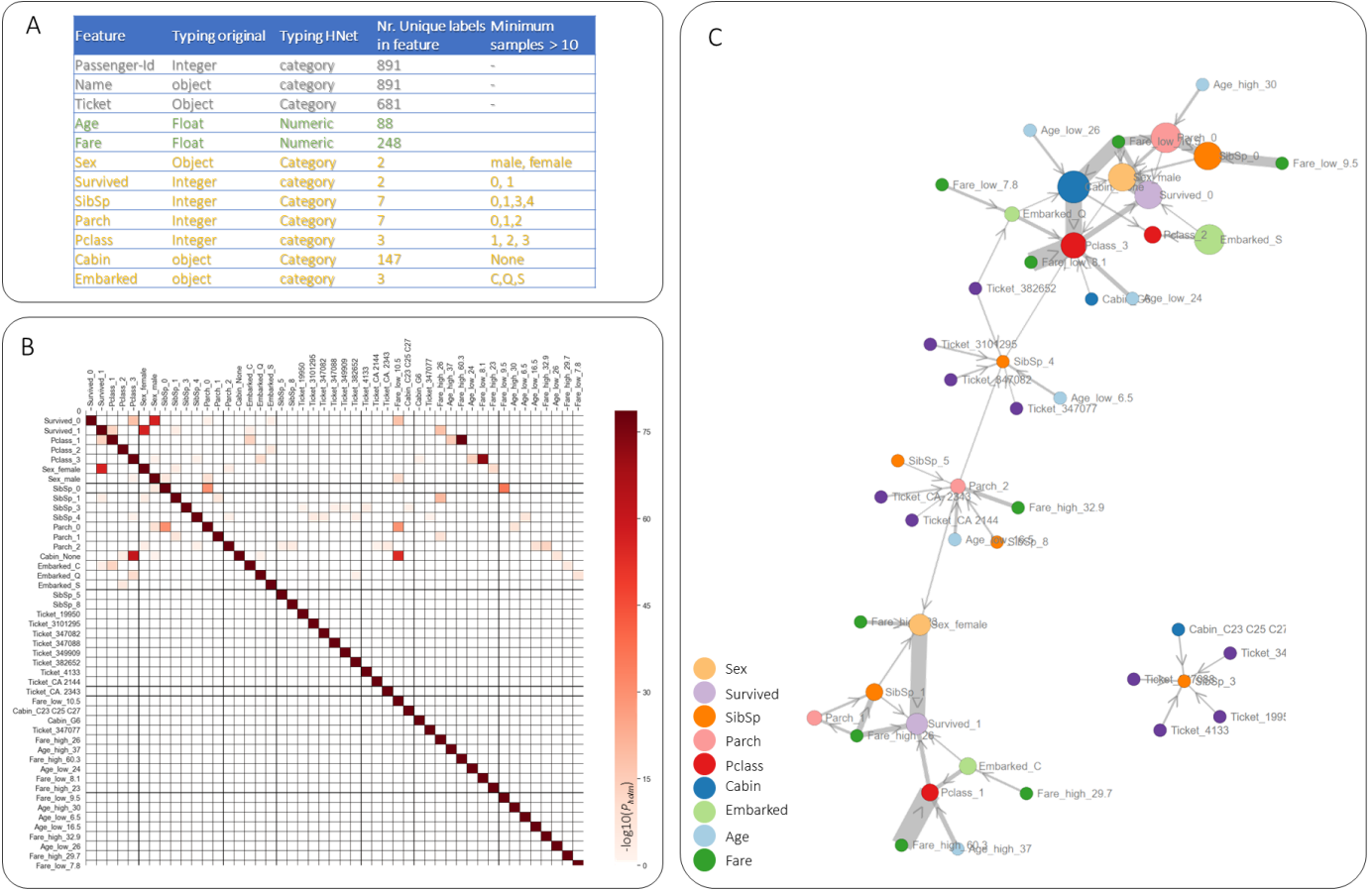

*Figure 4. **Results on Titanic data set using HNet.** (A) Table depicts the input features in the model, the original typing, the typing by HNet, the number of unique labels per feature, and the remaining labels that agree with the minimum of 10 samples. (B) Adjacency matrix determined by HNet for which the elements in the matrix are the node-links that are coloured on the -log10(Pholm). (C) Graph network view of the adjacency matrix. Node size is based on the percentage of available labels. Node colour is based on the unique feature names. Edge width between nodes is set by the -log10(Pholm) value.*

## DISCUSSION

Taken together, we demonstrate the detection of statistically significant associations in (un)structured data sets using HNet. The detected edges between nodes can be either directed or undirected. However, it should be noted that the directionality indicates that a feature is statistically overrepresented which does not necessarily imply causality. Furthermore, HNet provides deterministic results which can be deeper examined using the interactive network.

## AUTHOR CONTRIBUTIONS

ET designed the study, analysed the data, drafted the manuscript, and developed the python library.

## COMPETING INTERESTS

The author has declared that no competing interests exist.

*E.Taskesen*11. Chickering, D. M. Learning Bayesian Networks is NP-Complete. in (1996). doi:10.1007/978-1-4612-2404-4_12.
12. O'Gorman, B., Babbush, R., Perdomo-Ortiz, A., Aspuru-Guzik, A. & Smelyanskiy, V. Bayesian network structure learning using quantum annealing. *European Physical Journal: Special Topics* (2015) doi:10.1140/epjst/e2015-02349-9.
13. Agrawal, R. & Srikant, R. Fast Algorithms for Mining Association Rules in Large Databases. in *Proc. of the 20th International Conference on Very Large Data Bases (VLDB'94)* (1994).
14. Zaki, M. J. Scalable algorithms for association mining. *IEEE Trans. Knowl. Data Eng.* (2000) doi:10.1109/69.846291.
15. Martellini, M. & Malizia, A. *Cyber and Chemical, Biological, Radiological, Nuclear, Explosives Challenges: Threats and Counter Efforts*. (2017).
16. Beinlich, I. A., Suermondt, H. J., Chavez, R. M. & Cooper, G. F. The ALARM Monitoring System: A Case Study with two Probabilistic Inference Techniques for Belief Networks. in (1989). doi:10.1007/978-3-642-93437-7_28.
17. Singh, A., Saraswat, S. & Faujdar, N. Analyzing Titanic disaster using machine learning algorithms. in *Proceeding - IEEE International Conference on Computing, Communication and Automation, ICCCA 2017* (2017). doi:10.1109/CCAA.2017.8229835.
18. Benjamini, Y., Drai, D., Elmer, G., Kafkafi, N. & Golani, I. Controlling the false discovery rate in behavior genetics research. in *Behavioural Brain Research* (2001). doi:10.1016/S0166-4328(01)00297-2.
19. Hochberg, Y. A sharper bonferroni procedure for multiple tests of significance. *Biometrika* (1988) doi:10.1093/biomet/75.4.800.
20. Holm, S. A simple sequentially rejective multiple test procedure. *Scand. J. Stat.* (1979).
**6**